# *ROSE*: Real One-Stage Effort to Detect the Fingerprint Singular Point Based on Multi-scale Spatial Attention

Liaojun Pang, *Member, IEEE*, Jiong Chen, Fei Guo, Zhicheng Cao, and Heng Zhao

*Abstract*—Detecting the singular point accurately and efficiently is one of the most important tasks for fingerprint recognition. In recent years, deep learning has been gradually used in the fingerprint singular point detection. However, current deep learning-based singular point detection methods are either two-stage or multi-stage, which makes them time-consuming. More importantly, their detection accuracy is yet unsatisfactory, especially in the case of the low-quality fingerprint. In this paper, we make a Real One-Stage Effort to detect fingerprint singular points more accurately and efficiently, and therefore we name the proposed algorithm *ROSE* for short, in which the multi-scale spatial attention, the Gaussian heatmap and the variant of focal loss are applied together to achieve a higher detection rate. Experimental results on the datasets FVC2002 DB1 and NIST SD4 show that our *ROSE* outperforms the state-of-art algorithms in terms of detection rate, false alarm rate and detection speed.

## I. INTRODUCTION

THE fingerprint singular point, on which the directional field is not continuous, plays an important role in fingerprint classification, registration and indexing [1]. Although the fingerprint singular point detection has been studied for many years, it is still a challenge to detect the singular point accurately and efficiently from fingerprints, especially from *low-quality* fingerprints [2].
The early fingerprint singular point detection methods are mainly non-deep learning methods, including Poincaré Index (PI) methods [1, 3] and model-based methods [2, 4, 5], most of which can detect most singular points accurately from high-quality fingerprints. However, they require calculating the fingerprint orientation field before detection, which is time-consuming and they do not work very well for low-quality fingerprints.
  In recent years, with its outstanding performance in computer vision, deep learning has been gradually used in the fingerprint singular point detection, and several detection algorithms have been proposed based on deep learning. Jin *et al*. [6] firstly proposed a convolutional neural network (ConvNet) for estimating whether the center of a block is the singular point, and then ConvNet is transformed into a fully convolutional network (FCN) and fine-tuned. Unfortunately, their approach is somewhat complex because it has two networks to be trained and needs dividing each fingerprint image into blocks and classifying each block. Based on faster-RCNN [7], Liu *et al*. [8] proposed a two-stage network which consists of singular region proposals estimation and location regression. Although their method is simpler than Jin *et al*.'s [6], their two-stage detection process is still time-consuming, especially the region proposal network (RPN) [7].
  Geetika *et al*. [9] presented a fingerprint singular point detector which is made up of a macro-localization network and a micro-regression network and claimed that it is one-shot (i.e. one-stage). However, our thorough analyses show that Geetika *et al*.'s method is a not a real one-stage one in fact, because one-stage is a concept of detection algorithm which means one network and one training, which make it higher in efficiency compared to two-stage and multi-stage in general [10].
  In summary, with the help of deep learning technology, the methods mentioned above improve the detection performance, but experimental results show that they do not perform very well in the fingerprint singular point detection either, especially in the case of low-quality fingerprints. What's more, none of them is a real one-stage method. Motivated by these concerns, in this paper, we make a *Real One-Stage Effort* to detect fingerprint singular points based on the multi-scale spatial attention, and therefore name the proposed algorithm *ROSE* for short. Compared with all other algorithms aforementioned, the main innovations and contributions of this paper are as follows:
1. Based on the attention mechanism, we propose the multi-scale spatial attention in our algorithm *ROSE*, which improves the representation of singular points and focus attention on their positions in different scales.
2. We adopt the Gaussian heatmap instead of coordinate for labeling the ground truth, which could predict each pixel position and improve the positioning accuracy of the singular point. At the same time, we adopt the variant of focal loss as the loss function, which solves the problem to a certain extent that the proportion of positive and negative samples is unbalanced.
3. The proposed *ROSE* is a real one-stage method for

fingerprint singular point detection, which is a straightforward and efficient deep neural network (DNN) framework, relieved from the need of the Region Proposal Network (RPN). Such one-stage detection yields faster detection result.

## II. PROPOSED METHOD

The proposed method *ROSE*, a deep neural network, is a real one-stage algorithm for singular point detection and is composed of six types of operations including the basic spatial attention module, the convolution operation, the pooling operation, the upsampling operation, the multiply operation and the non-maximum suppression (NMS) operation.

### A. The overall framework of ROSE

As shown in Fig. 1, *ROSE* is constructed by three channels—the feature extraction channel, the core multi-scale spatial attention channel and the delta spatial multi-scale attention channel, followed by an extra NMS layer. The basic structure and function of each module is as follows:
(1) The feature extraction channel, which aims at extracting features from low-level to high-level, is composed of convolution operations and max-pooling operations.
(2) The core (or delta) multi-scale spatial attention channel, which is used to tell where we should pay attention and at the same time improve the representation of cores (or deltas) in different scales, is made up of the basic spatial attention and the upsamling operation. These two channels have the same structure.
(3) The NMS layer is at the end of *ROSE* and aims to remove redundant points and keep correct singular points simultaneously.

### B. The details of ROSE

With the fingerprint image as input, the feature extraction channel is made up of 10 convolution operations and 4 max-pooling operations, whose layout and operation sequence are shown in Fig. 1. All the ten convolution operations have a certain number of $3\times3$ filters with the same specifications, and the numbers of filters are 32, 32, 64, 64, 128, 128, 256, 256, 512 and 512, respectively. All the four max-pooling operations are set to window $2\times2$ and stride 2 (non-overlapping window). From top to bottom, the scale of tensors decreases gradually and the information becomes more abstract. The outputs of the feature extraction channel are used as the inputs of the core and delta multi-scale attention channels described below.

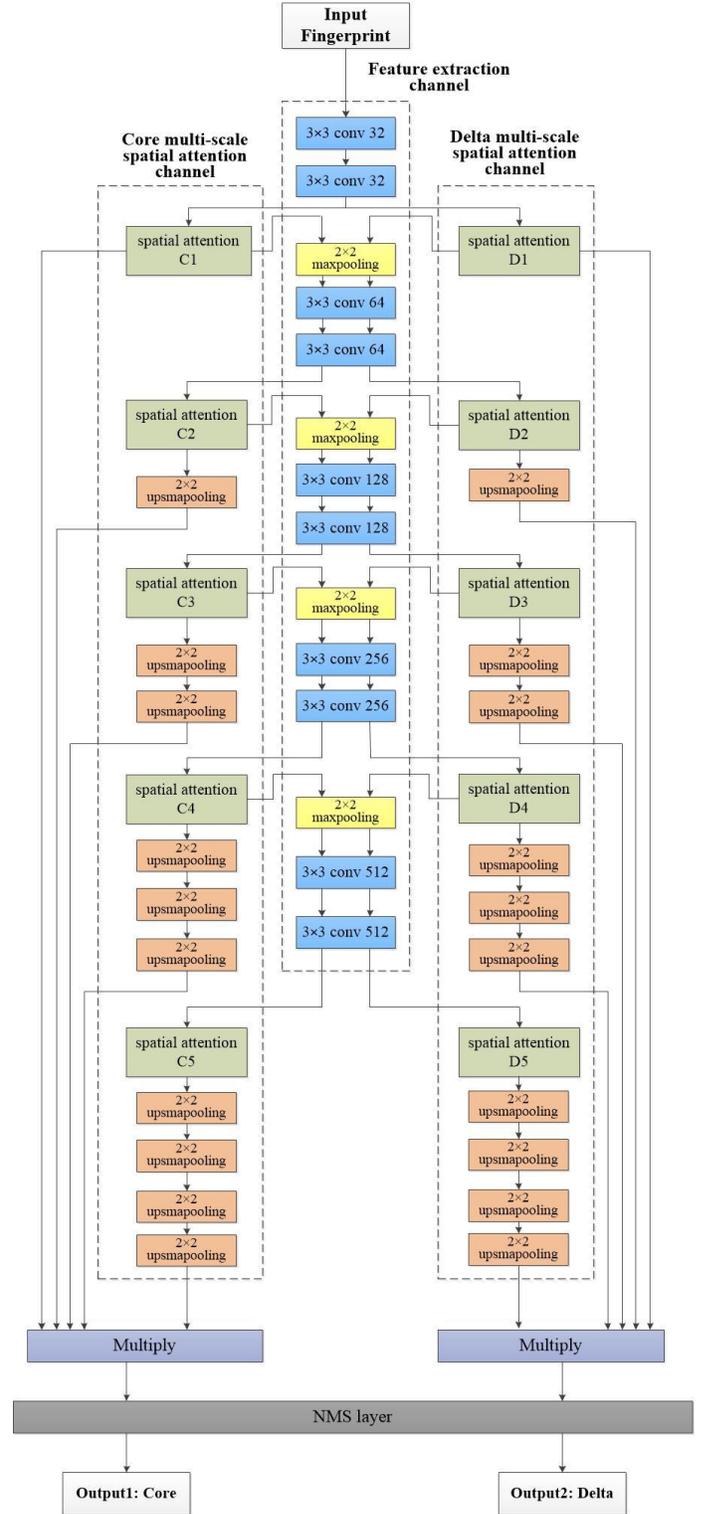

Fig. 1. The detailed *ROSE* structure

Because in our *ROSE*, the delta multi-scale attention channel has the same structure with the core one, we just take the latter for example to describe their structure. The core multi-scale attention channel consists of 5 basic spatial attention modules (from C1 to C5) and 10 upsampling operations with window $2\times2$. For C1 to C5, they have the same structure shown in Fig. 2. In a basic spatial attention module [11], with the feature maps from the feature extraction channel as input, two outputs are produced. One is the spatial attention map which tells where we should pay attention, and the other is the refined feature maps which improves the representation of cores. For the convolution operation used in the basic spatial attention module shown in Fig. 2, the filter number is 1, the kernel size is $5\times5$, the activation function is the sigmoid, and the stride is 1. Average-pooling and max-pooling operations are performed along the channel axis. It should be noted that the goal of the basic spatial attention module used in our *ROSE* is to increase the core representation power and suppress other unnecessary features.

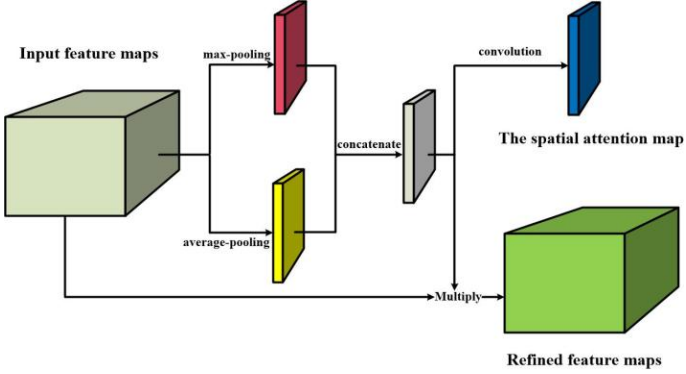

Fig. 2. The structure of a basic spatial attention module

Furthermore, for the core multi-scale spatial attention channel, as shown in Fig. 1, the scale of each input and output decreases by 0.5 times from C1 to C5, which constitutes multi-scale inputs and outputs. The refined feature maps output by C1 to C4 are also used as inputs for the 4 max-pooling operations of the feature extraction channel, respectively, and the output of C5 is just one spatial attention map. Behind the basic attention modules (C2 to C5), upsampling operations on each spatial map are carried out one by one from C2 to C5 and the scales of these maps are unified as the original input fingerprint image. In a word, the proposed multi-scale spatial attention channel combines the basic spatial attention module and the unsampling operation, which provides more information for the subsequent fusion results. As one of the most important contributions of our *ROSE* in fingerprint singular point detection, it not only tells where we should pay attention in different scales, but also improves the representation of cores or deltas from low-level to high-level.

The above is the description of the three channels, and the following is the output part of *ROSE*: the initial detection results of *ROSE* include two probability maps which are obtained by fusing the outputs from the core and delta multi-scale attention channels, respectively. For the core multi-scale attention channel, from C1 to C5, the five output attention maps of the same size are operated by the multiply function to obtain a fused probability map. Similarly, another fused probability map can be obtained by the delta multi-scale attention channel. These two maps are our initial detection results in which non-zero pixels represent the core or delta. Considering lots of non-zero points generated from our method, NMS is adopted to remove redundant points, whose neighborhood radius is set to 20 pixels and minimum value is set to 0.2. After that, the exact fingerprint singular point positions can be obtained.

### C. Labeling and objective function

The existing deep learning methods for the singular point detection employ coordinate as ground truth labels. In this paper, for the first time, the gaussian heatmap is adopted to make the ground truth label. Through the simulation of Gauss function, the closer the pixel is to the singular point, the bigger the value is. The probability of singular points is directly regressed by the gaussian heatmap. To a certain extent, each point provides supervision information, and the prediction of each pixel can improve the positioning accuracy [12].

In this paper, the objective function is defined as the variant focal loss which is mainly to solve the problem that the proportion of positive and negative samples in one-stage objective detection is seriously unbalanced. In general, the number of fingerprint singular points is limited, which results in few positive samples in the Gaussian heatmap. Hence, we think that the variant of focal loss [15] is suitable for the singular point detection. The formula of the variant of focal loss $L_{v\_focal\_loss}$ is as follows:

$$L_{v\_focal\_loss}\left(y,\hat{y}\right) = \frac{-1}{N}\sum_{j=1}^{h}\sum_{i=1}^{w}\begin{cases}\left(1-\hat{y}(i,j)\right)^2 \log\left(\hat{y}(i,j)\right) & \text{if } y(i,j)=1 \\ \left(1-y(i,j)\right)^4 \hat{y}(i,j)^2 \left(1-\log\left(\hat{y}(i,j)\right)\right) & \text{otherwise}\end{cases} \quad (1)$$

where $y$ is the true singular point heatmap, $\hat{y}$ is the predicted singular point heatmap, $N$ is the total number of targets in the map, $h$ is the height of the heatmap, $w$ is the width of the heatmap and $(i, j)$ is the coordinates of heatmap pixel.

As described above, *ROSE* has one simple and efficient deep neural network, one training and no RPN. One-time detection can obtain the detection result rapidly. Therefore, according to the concept of one-stage [10], *ROSE* is a real one-stage fingerprint singular point detection method.

## III. EXPERIMENTAL RESULTS

In this section, *ROSE* is evaluated on the public fingerprint datasets FVC2002 DB1 [13] and NIST SD04 [14], and then compared with four state-of-art singular point detection methods which include three deep learning methods [6, 8, 9] and one non-deep learning method [2]. The reason why we choose these two datasets lies in that these two different types of datasets are widely used in the evaluation of the singular point detection.

### A. Experimental Setup

FVC2002 DB1 contains 800 images and NIST SD04 contains 4000 images, some of which are low-quality ones. Half of fingerprint images from each dataset are randomly chosen as training data and the rest are used for testing. Due to lack of singular points location of fingerprint images in these two datasets, we have to manually calibrate all singular points for training and testing. In the training process, we choose Adam [16] as the optimization method. The learning rate is set to 0.01 and the momentum is set to 0.9. The experiment is carried out on a PC with Intel(R) Xeon (R) CPU E5-2630 (2.2 GHz), 128 GB of RAM, and GPU Tesla K40C.

*B. Singular Points Detection Performance*

In the testing process, the detection rate, the false alarm rate and the detection speed are chosen as the evaluation indicators and the evaluation criteria are the same as the Zhu *et al.*'s [2]. The total detection speed is defined as the average processing time.

With experiment results, a comparison is made between our *ROSE* and four state-of-the-art methods including walking point (WP) [2], block classification method (BC) [6], object detection (OD) [8] and SP-Net [9]. All the experiments are carried out under the similar conditions and follow the same evaluation criteria. The singular points detection accuracy of the five methods is shown in Table I and Table II, while the singular points detection speed in Table III.

From Table I and Table II, it can be seen that: (1) our *ROSE* is the best in the detection rate of both cores and deltas. Especially, our *ROSE* works much better than others in the detection rate of cores on NIST SD4, and it improves 3.6% compared to the best of the existing methods. (2) For the false alarm rate of both cores and deltas, our *ROSE* still works best. In a word, no matter which evaluation indicators considered *ROSE* performs best.

From Table III, we can see that the detection speed of our *ROSE* is 20 milliseconds per image averagely, which is much faster than other three deep learning methods [6, 8, 9] and even faster than the fastest non-deep learning method WP [2], which reflects the advantages of real one-stage of *ROSE*.

To further illustrate the advantages of our *ROSE* in detecting the singular points out of low-quality fingerprints, in Fig. 3, we present the detection results obtained by different methods on two low-quality fingerprint examples in the test dataset from NIST SD4. Fig. 3 (a) and (b) show the core detection results on F0318 and F0778. The yellow circle is the ground truth, the red circle is the result obtained by *ROSE*, the purple circle is the result obtained by OD, the blue circle is the result obtained by SP-Net, and the green circle is the result obtained by BC. As can be seen from Fig.3, all of OD, SP-Net and BC detect erroneous singular points, and WP and BC cannot even detect any point, while our *ROSE* can correctly detect the singular points. To some extent, it can be concluded that our *ROSE* is more effective and robust for detecting the singular points out of low-quality fingerprints than other existing methods.

TABLE I
PERFORMANCE OF THE FOUR SINGULAR POINTS DETECTION METHODS AND THE PROPOSED *ROSE* OVER THE FVC2002 DB1

| Algorithm | Detection rate (%) | | False alarm rate (%) | |
|---|---|---|---|---|
| | Cores | deltas | Cores | deltas |
| WP [2] | 94.8 | 97.8 | 0.9 | 4.2 |
| SP-Net [6] | 93.0 | 95.0 | 2.3 | 6.2 |
| BC [8] | 95.2 | 98.1 | 1.2 | 4.0 |
| OD [9] | 96.3 | 98.4 | 1.0 | 3.6 |
| Proposed *ROSE* | **97.1** | **98.6** | **0.8** | **3.2** |

TABLE II
PERFORMANCE OF THE FOUR SINGULAR POINTS DETECTION METHODS AND THE PROPOSED *ROSE* OVER THE NIST SD4

| Algorithm | Detection rate (%) | | False alarm rate (%) | |
|---|---|---|---|---|
| | Cores | deltas | Cores | deltas |
| WP [2] | 84.6 | 90.9 | 5.4 | 4.7 |
| SP-Net [6] | 86.3 | 92.3 | 7.1 | 6.3 |
| BC [8] | 88.2 | 94.5 | 5.8 | 4.2 |
| OD [9] | 89.9 | 94.7 | 4.9 | 3.3 |
| Proposed *ROSE* | **93.5** | **95.1** | **4.2** | **3.1** |

TABLE III
THE DETECTION SPEED OF THE FOUR SINGULAR POINTS DETECTION METHODS AND THE PROPOSED *ROSE*

| Algorithm | Ave. time (ms) |
|---|---|
| WP [2] | 28 |
| SP-Net [6] | 253 |
| BC [8] | 813 |
| OD [9] | 214 |
| Proposed *ROSE* | **20** |

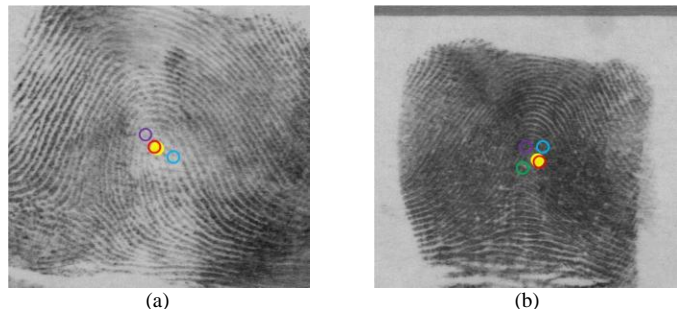

(a)     (b)
Fig. 3. Two examples of core detection results by the five singular points detection methods and the proposed *ROSE*

IV. CONCLUSION

For better accuracy and higher speed in the fingerprint singular point detection, we in this paper propose a new deep learning method, *ROSE,* based on multi-scale spatial attention. The results on FVC2002 DB1 and NIST SD4 indicate that: (1) the multi-scale spatial attention, the heatmap and the variant of focal loss ensure the high performance of the network which obtains higher detection rates as well as lower false alarm rates and (2) one-stage strategy *ROSE* works much more efficiently than two-stage or multi-stage strategies and even faster than non-deep learning methods. We think that the design idea and experimental advantages of *ROSE* may be helpful to other object detection researches based on deep learning.

REFERENCES

[1] J. Zhou, F. L. Chen, and J. W. Gu, "A novel algorithm for detecting singular points from fingerprint images," *IEEE Trans. Pattern Anal. Mach. Intell.*, vol. 31, no. 7, pp. 1239-1250, 2009.
[2] E. Zhu, X. F. Guo, and J. P. Yin, "Walking to singular points of fingerprints," *Pattern Recognit.*, vol. 56, pp. 116-128, 2016.
[3] D. W. Weng, Y. L. Yin, and D. Yang, "Singular points detection based on multi-resolution in fingerprint images," *Neurocomputing*, vol. 74, pp. 3376-3388, Oct 2011.
[4] L. L. Fan, S. G. Wang, H. F. Wang, and T. D. Guo, "Singular points detection based on zero-pole model in fingerprint images," *IEEE T Pattern Anal*, vol. 30, pp. 929-940, 2008.

[5] C. Jin and H. Kim, "Pixel-level singular point detection from multi-scale Gaussian filtered orientation field," *Pattern Recognit.*, vol. 43, pp. 3879-3890, Nov 2010.

[6] J. Qin, C. Han, C. Bai, and T. Guo, "Multi-scaling detection of singular points based on fully convolutional networks in fingerprint images," in *Proceedings of the Chinese Conference on Biometric Recognition*, pp. 221-230, 2017.

[7] S. Ren, K. He, R. Girshick, and J. Sun, "Faster R-CNN: Towards Real-Time Object Detection with Region Proposal Networks," in *Advances in neural information processing systems*, pp. 91-99, 2015.

[8] Y. H. Liu, B. C. Zhou, C. Y. Han, T. D. Guo, and J. Qin, "A Method for Singular Points Detection Based on Faster-RCNN," *Applied Sciences*, vol. 8, pp. 1853, no.10, 2018.

[9] G. Arora, R. R. Jha, A. Agrawal, K. Tiwari and A. Nigam, "SP-NET: One shot fingerprint singular-point detector," *arXiv preprint*, arXiv: 1908.04842, 2019.

[10] W. Liu, D. Anguelov, D. Erhan, C. Szegedy, S. Reed, C. Fu and A. C. Berg, " SSD: Single shot multibox detector," in *2016 European Conference on Computer Vision* (ECCV 2016), pp. 21-37, 2016.

[11] S. Woo, J. Park, J. Lee, and I. S. Kweon, "Cbam: Convolutional block attention module, " in *2018 Proceedings of the European Conference on Computer Vision* (ECCV 2018), pp. 3-19, 2018.

[12] G. Papandreou, T. Zhu, N. Kanazawa, A. Toshev, J. Tompson, C. Bregler and K. Murphy, "Towards Accurate Multi-Person Pose Estimation in the Wild," in *2017 IEEE Conference on Computer Vision and Pattern Recognition* (CVPR 2017), pp. 4903-4911, 2017.

[13] D. Maio, D. Maltoni, R. Cappelli, J. L. Wayman and A.K. Jain, "FVC2002: Second fingerprint verification competition," in *Object recognition supported by user interaction for service robots*, vol. 3, 2002.

[14] C. I. Watson and C. L. Wilson, "NIST Special Database 4, Fingerprint Database," National Institute of Standards and Technology, 1992.

[15] H. Law and J. Deng, "CornerNet: Detecting Objects as Paired Keypoints," in *2018 Proceedings of the European Conference on Computer Vision* (ECCV 2018), pp. 734-750, 2018.

[16] P. K. Diederik, and B. L. Jimmy, "Adam: A Method for Stochastic Optimization," *arXiv preprint*, arXiv: 1412.6980, 2014.